\documentclass[aps,prap,reprint,groupedaddress]{revtex4-2}

\usepackage{xcolor}
\usepackage{graphicx}
\usepackage{dcolumn}
\usepackage{bm}
\usepackage{amsmath}

\newcommand\jlin  [1]{{\color{black} #1}}

\begin{document}

\title{Optimal swimming with body compliance in an overdamped medium}



\author{Jianfeng Lin $^{1,2,\dagger}$}
\author{Tianyu Wang $^{1,3,4,\dagger}$}
\author{Baxi Chong $^5$}
\author{Matthew Fernandez $^{3}$}
\author{Zhaochen Xu$^{1,2}$}
\author{Daniel I. Goldman$^{1}$}
\email{Corresponding author: daniel.goldman@physics.gatech.edu}  

\affiliation{$^\dagger$These authors contributed equally to this work.}

\affiliation{$^1$School of Physics, Georgia Institute of Technology, Atlanta, GA, USA}
\affiliation{$^2$Interdisciplinary Graduate Program in Quantitative Biosciences, Georgia Institute of Technology, Atlanta, GA, USA.}
\affiliation{$^3$George W. Woodruff School of Mechanical Engineering, Georgia Institute of Technology, Atlanta, GA, USA }
\affiliation{$^4$Institute for Robotics and Intelligent Machines, Georgia Institute of Technology, Atlanta, GA, USA }
\affiliation{$^5$Department of Mechanical Engineering, The Pennsylvania State University, University Park, PA, USA}



\date{\today}

\begin{abstract}

Elongate animals and robots use undulatory body waves to locomote through diverse environments. Geometric mechanics provides a framework to model and optimize such systems in highly damped environments, connecting a prescribed shape change pattern (gait) with locomotion displacement. However, the practical applicability of controlling compliant physical robots remains to be demonstrated. In this work, we develop a framework based on geometric mechanics to predict locomotor performance and search for optimal swimming strategies of compliant swimmers. We introduce a compliant extension of Purcell's three-link swimmer by incorporating series-connected springs at the joints. Body dynamics are derived using resistive force theory. Geometric mechanics is incorporated into movement prediction and into an optimization framework that identifies strategies for controlling compliant swimmers to achieve maximal displacement. We validate our framework on a physical cable-driven three-link limbless robot and demonstrate accurate prediction and optimization of locomotor performance under varied programmed, state-dependent compliance in a granular medium. Our results establish a systematic, physics-based approach for modeling and controlling compliant swimming locomotion, highlighting compliance as a design feature that can be exploited for robust movement in both homogeneous and heterogeneous environments.

\end{abstract}

\maketitle


\section{Introduction}

\begin{figure}[t]
\centering
\includegraphics[width=1\columnwidth]{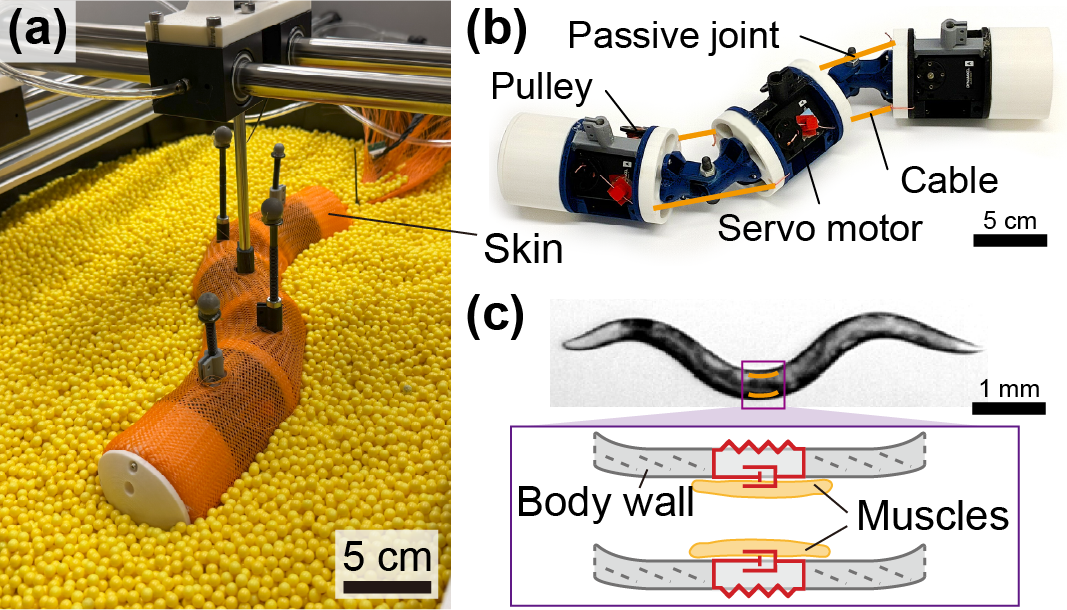}
\caption{\textbf{Concept of body compliance in a cable-driven three-link robophysical model and nematodes \textit{C. elegans}. } (a) The three-link cable-driven swimmer, mounted on a gantry, was immersed in a granular medium. (b) Three-link cable-driven swimmer robophysical model (skin off), with bilateral cables routed through pulleys and actuated by servo motors to produce in-plane bending and body compliance. (c) The nematode \textit{C. elegans} has the compliant body walls and bilateral muscles along its body.}
\label{fig:intro}
\end{figure}

Undulatory locomotion, in which traveling body waves generate self-propulsion, is a fundamental strategy employed by elongate animals (e.g., snakes~\cite{gray1946mechanism,hu2009mechanics,jayne1986kinematics,astley2015modulation,tingle2024functional} eels~\cite{gillisEnvironmentalEffectsUndulatory1998,tytell_hydrodynamics_2004}, skinks~\cite{maladen2009undulatory,chongCoordinatingTinyLimbs2022}) and robots~\cite{hirose2004biologically,transeth2008snake,crespi2008online,wright2007design,crespi2013salamandra,chong2021frequency,maladen2011undulatory} to traverse both fluid and terrestrial environments. These locomotive capabilities emerge from the interplay between body dynamics and environmental media~\cite{guo2008limbless,aguilar2016review}, motivating extensive research efforts aimed at uncovering substantial principles and deriving unified frameworks for understanding animal movements and informing robotic developments~\cite{guo2008limbless,hatton2013geometric,maladen2009undulatory,rieser2024geometric,chong2021frequency,charles2025topological}.


Recent studies focusing on highly damped environments (e.g., viscous fluids~\cite{fu2007theory}, granular media~\cite{marvi2014sidewinding,maladen2009undulatory}, and frictional substrates~\cite{chong2023self}), where inertial effects can be neglected, have shown that undulatory locomotion can be effectively modeled within a general geometric framework known as geometric mechanics (also referred to as the geometric phase approach)~\cite{hatton2013geometric,rieser2024geometric}. The concept of geometric phase was originally developed in quantum mechanics~\cite{berry1988geometric} and later applied to classical mechanics~\cite{marsden1990reduction}. Specifically, Shapere and Wilczek demonstrated that the geometric phase approach could provide important insights into the locomotion of microorganisms~\cite{shapere1987self}. \jlin{Further, beginning in the 1980s, physicists and control theorists extended the geometric phase framework to a variety of systems moving through different environments. For example, beginning with Purcell's three-link swimmer~\cite{purcell1977life}, a simple model used to study locomotion in low–Reynolds-number fluids~\cite{becker2003self,tam2007optimal}, geometric mechanics has been extended to viscous fluids~\cite{hatton2013geometric_tro,ramasamy2019geometry} and dry granular media~\cite{hatton2013geometric}, where the dynamics can be determined analytically and empirically. More broadly, with appropriate dimension-reduction techniques, geometric mechanics has proven effective in modeling animal locomotion across diverse environments~\cite{rieser2024geometric} and in guiding the control of limbless robots~\cite{chong2021frequency,chong2023optimizing,ramasamy2019geometry}.}



Though effective in modeling undulatory locomotion in highly damped environments, geometric mechanics tools are limited to linking \textit{determined} self-deformation patterns (gaits) with their resulting locomotor performance~\cite{shapere1987self}. However, in the practical control of limbless robots or the neuromechanics of elongate organisms, shape-change patterns are not directly prescribed but instead emerge from interactions between body viscoelasticity and the surrounding homogeneous or heterogeneous environment. This interaction-involved “loop” plays a central role in the gait adaptation of \textit{C. elegans}~\cite{pierce2024dispersion,pierce2025neuromechanical} (Fig.~\ref{fig:intro}(c)) and has been shown to simplify robot control in rheologically complex media~\cite{wang2023mechanical}. \jlin{To address this, Ramasamy and Hatton~\cite{ramasamy2021optimal} extended the geometric framework to identify optimal gaits for compliant swimmers. The framework successfully modeled and optimized performance under system constraints for two types of proposed Purcell’s three-link swimmers, each with one actuated joint and one purely elastic joint, in a viscous fluid (similar to the systems described in~\cite{zigelman2024dynamics}). However, its broader applicability to physical robotic platforms remains to be demonstrated.}

Similar attention has been devoted in robotics to the optimal control of systems with elastic actuation~\cite{pratt1995series,vanderborght2013variable,chhatoi2023optimal,yasa2023overview,wang2020directional}. Inspired by the compliance observed in biological organisms, elastic actuation---achieved through actuators incorporating elastic components or soft materials---has been widely implemented in manipulators~\cite{lin2024bioinspired,wang2018programmable,mura2018soft}, legged robots~\cite{hutter2016anymal,chong2023multilegged,lin2025bird,badri2022birdbot}, and exoskeletons~\cite{qian2022curer,cherry2016running,collins2015reducing} to mitigate external perturbations, for safer physical interactions and/or improved adaptability to complex environments. However, the added compliance complicates control by introducing nonlinear dynamics and frequency-dependent behavior, particularly in contact-rich tasks. As a result, controlling such systems often requires computationally intensive methods, which can limit their practical deployment in real-world applications~\cite{schumacher2019introductory,alessi2024rod,haggerty2023control}.

\jlin{In this work, we develop a framework based on geometric mechanics to model and optimize the performance of compliant swimmers in overdamped environments. To achieve this, we introduce a compliant extension of Purcell's three-link swimmer by incorporating series-connected springs at both joints (as shown in Fig.~\ref{fig:springModel}), in which the realized gait emerges from interactions between actuation patterns, body compliance and the environment. We first model the system dynamics using resistive force theory (RFT) and employ geometric mechanics tools to predict locomotor performance.} We then propose a gait optimization framework that identifies gaits that yield maximal displacement from geometric mechanics and generates their corresponding control commands through inverse body dynamics. We validate the framework by applying it to a physical cable-driven limbless robot capable of exhibiting tunable, state-dependent compliance (Fig.~\ref{fig:intro}(a) and (b)) and demonstrate that it can accurately predict and optimize its locomotion under varied body compliance in a granular medium.

\section{Method}

\subsection{System dynamics}

\begin{figure}[t]
\centering
\includegraphics[scale=.9]{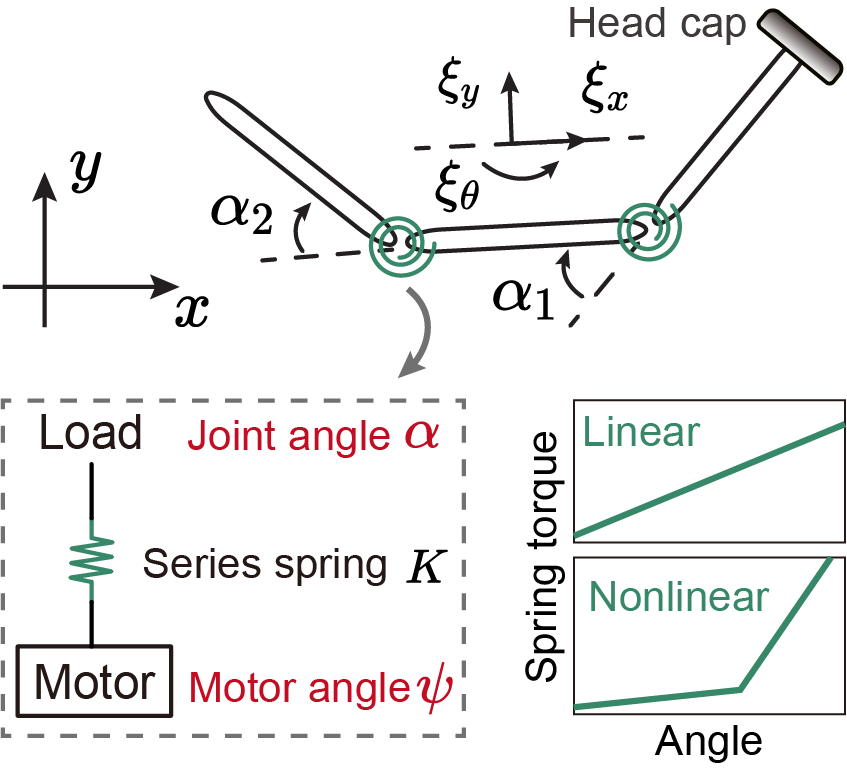}
\caption{\textbf{Analytical model of the compliant three-link swimmer.} Analytical three-link model with a body frame corresponding to a weighted average of the link positions and orientations. Each joint includes a motor connected in series with a spring. Insets illustrate both linear and nonlinear springs, which can be captured by the model.}
\label{fig:springModel}
\end{figure}

The model is based on Purcell's classic three-link swimmer model, comprising three rigid segments connected in series. Body compliance is incorporated by introducing a spring in series with each motor as illustrated in Fig.~\ref{fig:springModel}. For locomotion in highly damped environments (also referred to as low-coasting environments~\cite{rieser2024geometric}), where inertial effects are typically negligible, environmental forces are assumed to be balanced by internal body compliance (spring forces). Under these assumptions, the governing equations describing system dynamics are:
\begin{equation}
\underbrace{\bm{K}(\bm{\alpha})(\bm{\psi}-\bm{\alpha})}_{\text{body torque}}=\bm{\tau}_{\rm{env}}(\bm{\alpha},\dot{\bm{\alpha}}),
\label{eq:dyn}
\end{equation}
where $\bm{\alpha} = [\alpha_1, \alpha_2]^{\rm{T}}$ represents the emergent shape of the swimmer (joint angles that the swimmer realize, hereafter we call $\alpha$ as ``emergent" joint angles); $\bm{\psi}$ denotes the shape we command the swimmer to form by actuating the cables, hereafter we call $\psi$ as ``suggested" joint angles; $\bm{K}(\bm{\alpha})$ is the $2 \times 2$ joint stiffness matrix; and $\bm{\tau}_{\rm{env}}(\bm{\alpha},\dot{\bm{\alpha}})$ represents environmental torques acting on each joint. In this study, the joint stiffness can either be constant or state-dependent explicitly on the actual joint angles as shown in Fig.~\ref{fig:springModel}.

\begin{figure}[t]
\centering
\includegraphics[scale=.9]{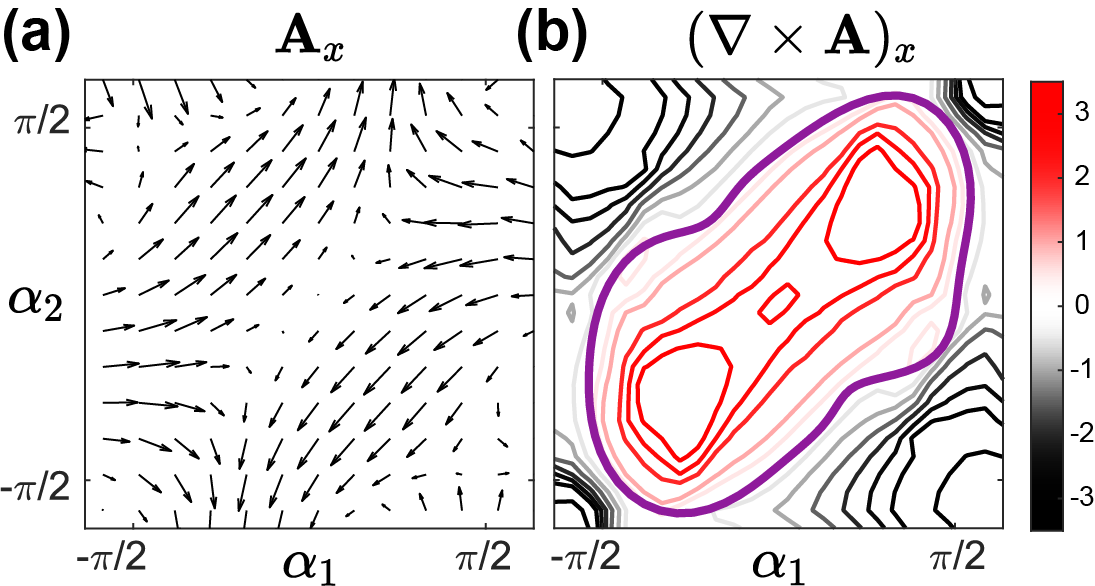}
\caption{\textbf{Tools from geometric mechanics for modeling and optimization.} (a) The local connection vector field from geometric mechanics, which map joint velocities to body velocities and provide the foundation for displacement prediction. (b) The height function, the curl of the local connection vector field. The net displacement from a gait (purple) corresponds to the areas it encloses on the height function. The unit of the height function is (body length)/(radian$^2$), and its values are scaled by a factor of 100.
}

\label{fig:HF}
\end{figure}

The environmental forces acting on each link are modeled using resistive force theory (RFT), which decomposes forces into local normal and tangential components relative to the link surface. We consider two distinct highly damped environments: a viscous fluid and a 6-mm granular medium. In viscous fluids, resistive forces are proportional to link velocities, with a higher drag coefficient in the normal direction than the tangential direction—a phenomenon known as drag anisotropy. Similarly, locomotion within granular media is captured using granular resistive force theory (granular RFT), an effective rate-independent force model that accounts for the dependence of forces on velocity direction and has successfully described undulatory locomotion in granular media~\cite{aguilar2016review}. To more accurately represent the real-world robophysical system, introduced in the following section, we model the body as a series of connected links without accounting for width, while incorporating a ``T"-shaped element at the first link to represent the head as shown in Fig.~\ref{fig:springModel}.

\subsection{Geometric Mechanics}

The system dynamics for a given motor command are solved based on equation (\ref{eq:dyn}). In highly damped environments where the locomotion is approximated  as quasi-static, the kinematics can be estimated as
\begin{equation}
\bm{\xi}=\bm{A}(\bm{\alpha})\dot{\bm{\alpha}}= 
\begin{bmatrix}
        \bm{A}_x(\bm{\alpha}) \\ \bm{A}_y(\bm{\alpha}) \\ \bm{A}_\theta (\bm{\alpha})
\end{bmatrix}\dot{\bm{\alpha}},
\label{eq:gmequ}
\end{equation}
where $\bm{\xi} = [\xi_x, \xi_y,\xi_{\theta}]^{\rm{T}}$ refers to the body velocity in forward, lateral and rotational directions as shown in Fig.~\ref{fig:HF}(a), $\bm{\alpha}$ is the actual joint angles as mentioned before, and $\bm{A}(\bm{\alpha})$ is the $3 \times 2$ local connection matrix which maps the joint angular velocity with body velocity linearly. Each row of matrix represents the direction of body velocity and can be visualized as a connection vector field in shape space as shown in Fig.~\ref{fig:HF}. For viscous fluid or granular media, the local connection matrix can be analytically or numerically solved by the following force $F_{x,y}$ and torque $M$ balance.
\begin{equation}
\sum_{i}\bm{F}_i(\bm{\xi},\bm{\alpha},\bm{\dot{\alpha}}) = \sum_{i}[F_{x}^i,F_{y}^i,M_i]^{\rm{T}}=0,
\label{eq:forceIntegral}
\end{equation}
where $\bm{F}_i$ denotes the environmental force applied on each link. With the local connection matrix, the body velocity can be calculated as the product of the connection matrix and joint angular velocity. 

For a periodic gait $\{\partial \phi\ : [\alpha_1(t), \alpha_2(t)], t\in(0,T]\}$, where $T$ is the temporal period, the displacement in x, y, and rotational directions ($\Delta x,\Delta y,\Delta \theta$) can be approximated to the first order as:
\begin{equation}
        \begin{pmatrix} 
        \Delta x \\
        \Delta y \\
        \Delta \theta 
    \end{pmatrix}\approx \int_{\partial \phi}\bm{A}(\bm{\alpha})\dot{\bm{\alpha}}.
\label{eq:lineIntergral}
\end{equation}
Further, according to Stokes' Theorem, the line integral along a closed curve $\partial \phi$ is equal to the surface integral of the curl of $\bm{A}(\bm{\alpha})$ over the surface enclosed by the path. Equation (\ref{eq:lineIntergral}) can be derived as

\begin{equation}
\begin{pmatrix} 
        \Delta x \\
        \Delta y \\
        \Delta \theta 
    \end{pmatrix} \approx \iint_{\phi}\nabla \times \bm{A}(\bm{\alpha} )\rm{d}\alpha_1\rm{d}\alpha_2,
\label{eq:heightFunc}
\end{equation}
where $\phi$ is the enclosed surface by path $\partial \psi$. $\nabla \times \bm{A}(\bm{\alpha})$ refers to the height function. Specifically, $(\nabla \times \bm{A})_x$ illustrated in Fig.~\ref{fig:HF}(b) presents the forward height function, which quantifies the net forward placement for one completed gait cycle.

With the height function, the gait search problem now is to draw the closed gait path over height function in shape space. Optimal displacement gait can be visually identified in the height function where the closed path maximizes the surface integral. For example, the optimal forward displacement gait in a granular medium of 6-mm diameter plastic spheres is shown as the purple path in Fig.~\ref{fig:HF}(b). Practically, gait execution is constrained by the geometric limits of the physical robot. In our experiments, each joint is limited to a range of 75$^\circ$. When the calculated optimal gait exceeds this range, the values are truncated to the joint limit.

\subsection{Gait optimization flow}

\begin{figure}[t]
\centering
\includegraphics[scale=.9]{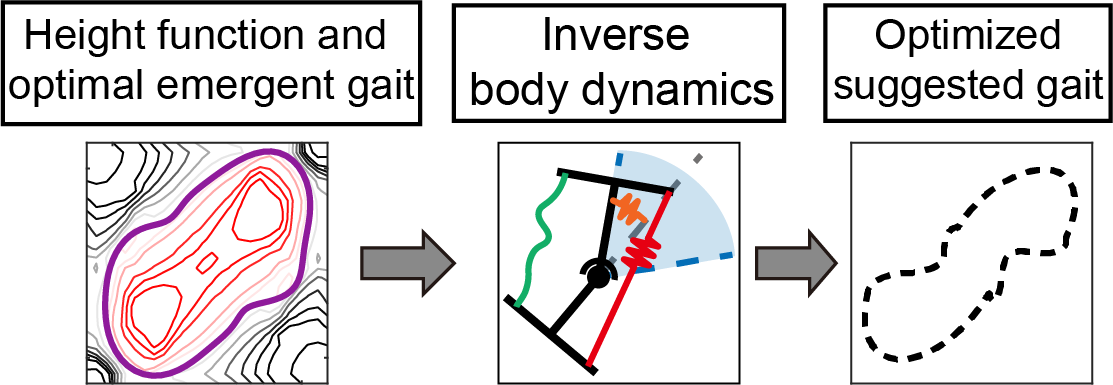}
\caption{\textbf{Optimization flow for identifying optimal gaits under body compliance.} First, the optimal emergent gait is identified by deriving the height function. Then, by incorporating the inverse body dynamics, the corresponding optimized gait is obtained.
}
\label{fig:OptiFlow}
\end{figure}

The goal for this section is to search the optimal joint angle sequence in the motor space ($\bm{\psi}$, the ``suggested" joint angle) to have the swimmer emerge an joint angle sequence in shape space (\textbf{$\alpha$}, the ``emergent" joint angle) that generates the maximum forward displacement under body-environment interactions. We prescribed the commanded motor angles $\psi_i$ for two joints ($i={1,2}$) motor space as 10th-order Fourier series
w\begin{equation}
    \psi_i=\sum_{p=1}^{10} a_p^i\cos (\frac{2\pi pt}{T} )+b_p^i\sin (\frac{2\pi pt}{T} ),
\end{equation}
where $T$ is the temporal period of gait. The optimization flow is described in Fig.~\ref{fig:OptiFlow}. To simplify the calculation we incorporate the height function from geometric mechanics. After deriving the height function from the certain medium (viscous fluid or granular media), the optimal gait can be easily identified from the height function. With the optimal gait in shape space, we can directly solve the motor space parameters $\{a_1^i,...,a_{10}^i,b_1^i,...,b_{10}^i\}$ from Equation (\ref{eq:dyn}).

\section{Robophysical implementation}
\subsection{Robophysical Model Development}
The robophysical model used in this study is a three-link (two-joint), cable-driven limbless swimmer. Both joints allow in-plane rotational bending, enabling the swimmer to execute planar undulatory motions. All three modules are mechanically identical, each with a length of 10 cm and diameter of 7.5 cm. Each module consists of a 3D-printed PLA outer shell that houses one DYNAMIXEL 2XL430-W250-T (ROBOTIS) actuator. This actuator contains two independently controlled servo motors, each driving a pulley with a 9.5 mm inner diameter. Non-elastic cables are spooled around the pulleys to form antagonistic pairs on two sides of the joint. These cables exhibit negligible shape memory and minimal elongation under load, ensuring accurate and consistent transmission of actuation. The distal ends of the cables are anchored to the adjacent module, enabling joint motion via differential cable retraction. The swimmer is wrapped in a mesh skin made of 4 cm inner diameter expandable sleeving (McMaster-Carr). This outer layer smooths the body profile and prevents granular particles from entering the joint gaps, which could otherwise cause jamming or hindered motion.

\begin{figure}[t]
\centering
\includegraphics[width=0.9\columnwidth]{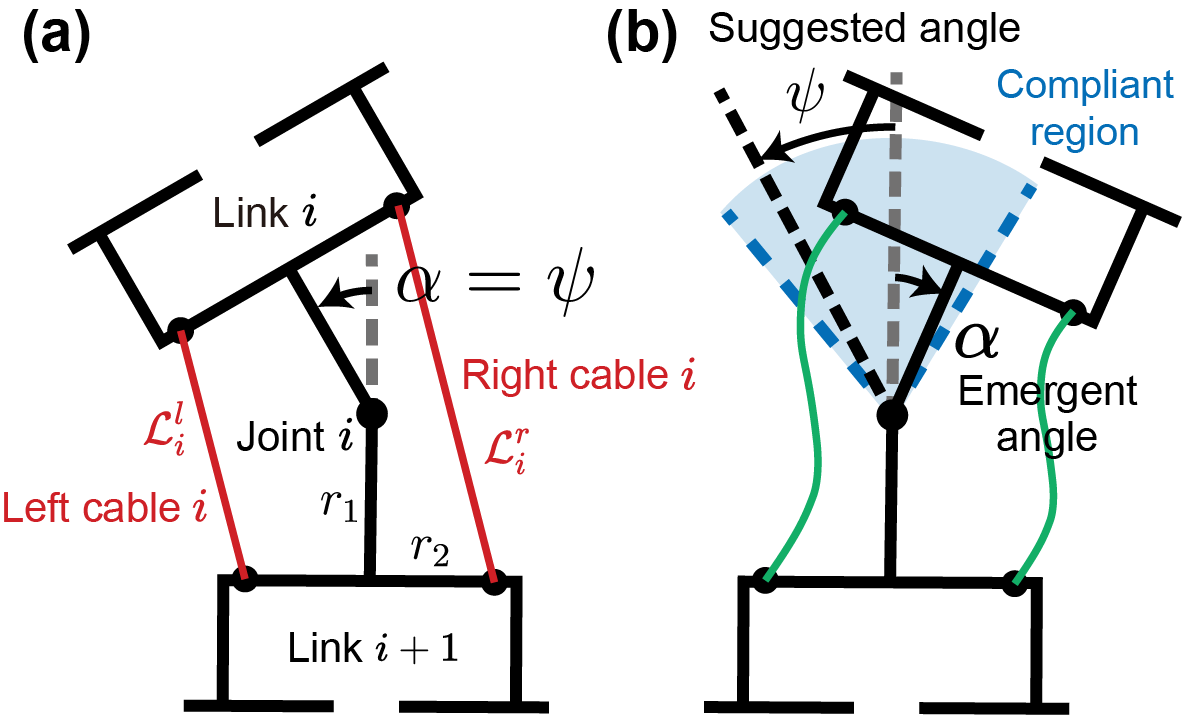}
\caption{\textbf{Cable-driven robophysical model design and joint compliance mechanism.} Schematic of bilateral cable actuation at a single joint, where left and right cables tensioned to form the exact suggested joint angle (b), and left and right cable slacked to form a compliant region, so that the emergent joint angle can deviate from the suggested angle (c).}
\label{fig:robot}
\end{figure}

\subsection{Cable driving scheme}

\begin{figure}[t]
\centering
\includegraphics[width=0.9\columnwidth]{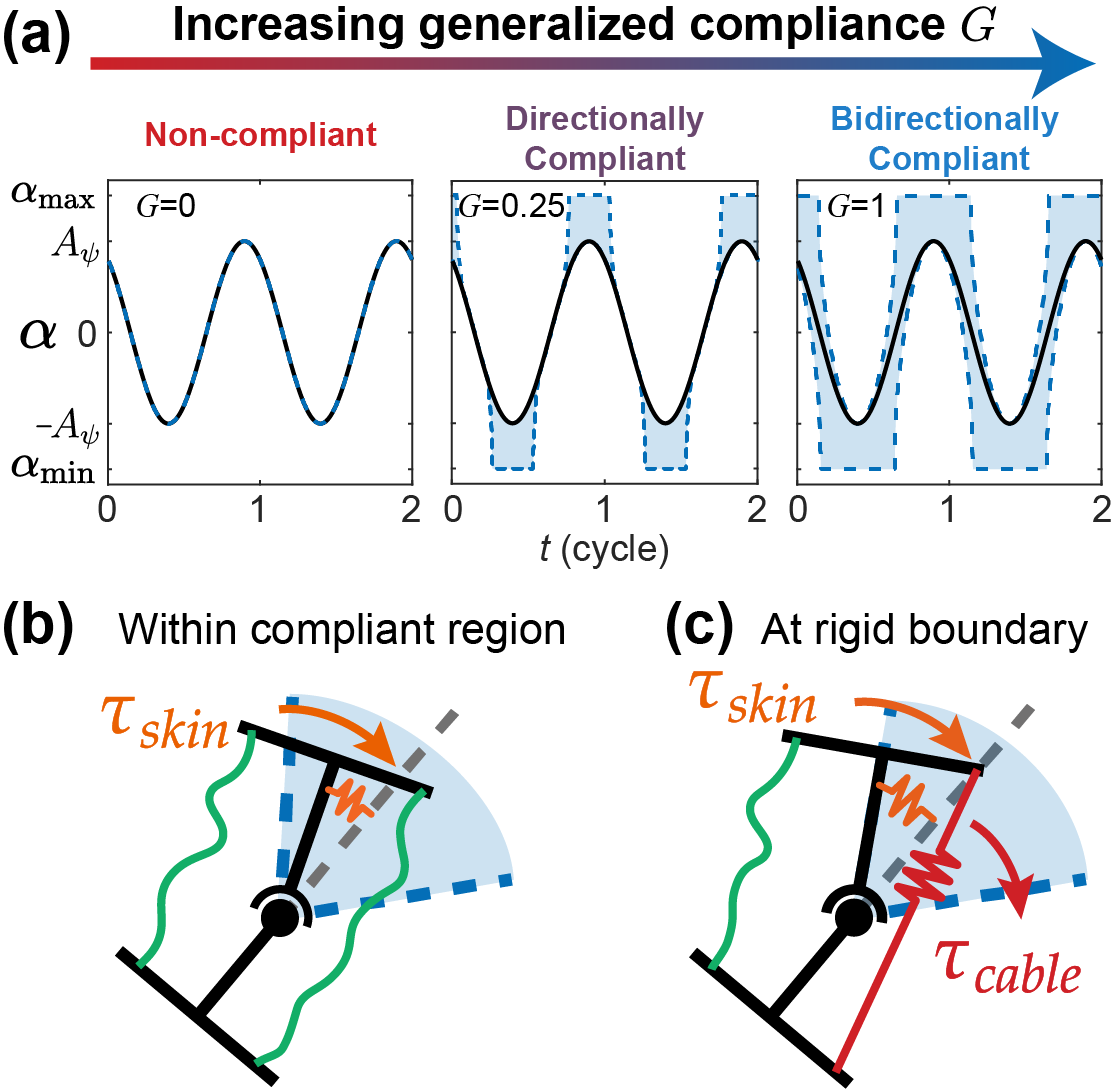}
\caption{\textbf{Effect of generalized compliance $G$ on joint behavior.} (a) Emergent joint angle trajectories ($\alpha$) across different compliance regimes: rigid (non-compliant, $G=0$), directionally compliant ($G=0.25$), and bidirectionally compliant ($G=1$). Compliance enlarges the range of joint motion within which the suggested angle ($\psi$) can deviate, illustrated by shaded blue regions. (b) Within the compliant region, the joint is governed primarily by the skin’s restoring torque ($\tau_\text{skin}$). (c) At the rigid boundary, cable tension engages, introducing an additional restoring torque ($\tau_\text{cable}$) that stiffens the joint response.}
\label{fig:G}
\end{figure}

To verify the modeling and optimization scheme, we introduce body compliance in the swimmer through a bilateral cable-driven scheme, which has been shown to facilitate body compliance in undulatory locomotion and thereby simplify locomotion control~\cite{wang2023mechanical}. Specifically, each joint is actuated by two cables, whose lengths can be independently controlled as $L_i^r$ and $L_i^l$ for the right and left cables, respectively, with the geometric relationship illustrated in Fig.~\ref{fig:G}(a). We define a generalized compliance parameter $G$ to control the simultaneous contraction of cables at each joint, governed by the following equations:

\begin{equation}
\begin{array}{l}
L_{i}^{l}\left(\psi_{i}\right)=\left\{\begin{array}{ll}
\mathcal{L}_{i}^{l}\left(\psi_{i}\right), & \text { if } \psi_{i} \leq-\gamma \\
\mathcal{L}_{i}^{l}[-\min (A_{\psi}, \gamma)]+l_{0} \left[\gamma+\psi_{i}\right] & \text { if } \psi_{i}>-\gamma
\end{array}\right. \\
L_{i}^{r}\left(\psi_{i}\right)=\left\{\begin{array}{ll}
\mathcal{L}_{i}^{r}\left(\psi_{i}\right), & \text { if } \psi_{i} \geq \gamma \\
\mathcal{L}_{i}^{r}[\min (A_{\psi}, \gamma)]+l_{0} \left[\gamma-\psi_{i}\right] & \text { if } \psi_{i}<\gamma
\end{array}\right.
\end{array},
\label{eq:gcable}
\end{equation}
where $\mathcal{L}_i^l$ and $\mathcal{L}_i^r$ denote the exact lengths of the left and right cables corresponding to the suggested motor angles, based on the geometry shown in Fig.~\ref{fig:G}(a). Here, $A{\psi}$ is the commanded gait amplitude, $l_0$ is a fixed design parameter related to cable tightness, and $\gamma=(2G-1)A{\psi}$.

Within this scheme, by varying the generalized compliance $G$, one cable becomes loose while the other is tightened, as illustrated in Fig.~\ref{fig:G}(b). The loose cable defines a compliant region of the joint that can rotate freely. Depending on $G$, the joint exhibits different behaviors:

(1) a rigid joint when $G=0$;

(2) a directionally compliant joint (e.g., $G=0.5$), where the joint can bend in one direction but not the other;

(3) a bidirectionally compliant joint (e.g., $G=1$), where the joint can bend in both directions.

With the generalized compliance $G$, the equivalent joint stiffness becomes state dependent. When the actual joint angle remains within the compliant region, the swimmer’s skin acts as a weak spring, restoring the joint toward the zero position. When the joint angle reaches the rigid boundary, the cable tightens and introduces an additional spring in series. The state-dependent body compliance can therefore be expressed as

\begin{equation}
K(a)=\begin{cases}k_{\text{skin} }
  & \text{ if } \alpha\in \text{compliant region} 
\\ k_{\text{skin}}+k_{\text{cable} }
& \text{ if } \alpha\in \text{rigid boundary} 
\end{cases},
\label{eq:stiffness}
\end{equation}
where $k_{\text{skin}}$ and $k_{\text{cable}}$ are the stiffness values of the skin and the cable, respectively, as measured in previous work~\cite{wang2023mechanical}.

\section{Experiments}
\begin{figure}[t]
\centering
\includegraphics[scale=.9]{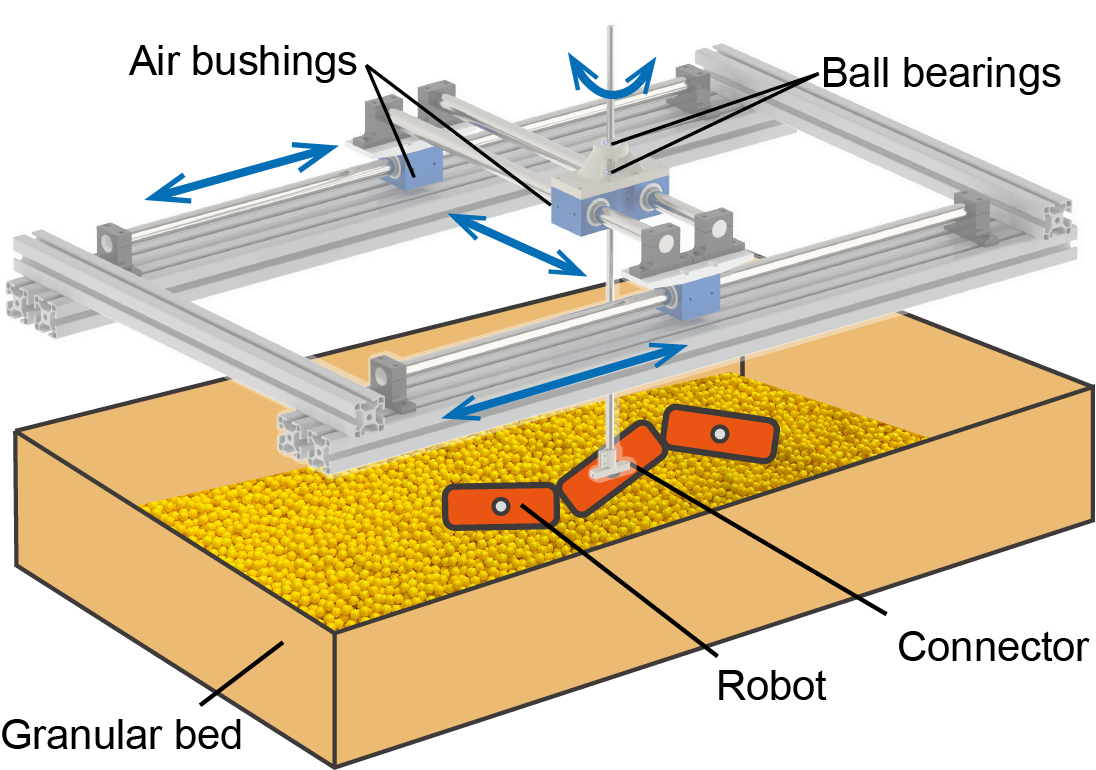}
\caption{\textbf{Experimental setup for robophysical experiments in granular media.} The robophysical model is immersed in granular media and mounted on a gantry, which constrains its motion to the horizontal plane while allowing both translation and rotation.
}
\label{fig:ExpSet}
\end{figure}

Experiments were conducted in a testbed filled with plastic spheres (6 mm diameter), which served as the granular environment. To constrain the swimmer's motion to a horizontal plane and maintain a consistent depth, we designed a custom gantry system incorporating air bearings. The system includes two orthogonal linear air bearings (New Way Air Bearings) that provide frictionless translation along the X and Y axes, and a rotational ball bearing enabling smooth in-plane rotation as shown in Fig.~\ref{fig:ExpSet}. The swimmer is attached to the gantry via a vertical shaft whose lower end is fixed to a rigid connector mounted at the center of the middle link. This mechanical coupling ensures that the swimmer remains suspended at a constant height while allowing free planar motion. The overall setup minimizes mechanical resistance, eliminates vertical drag or sinkage, and enables repeatable locomotion trials within the granular bed.

To track the motion of the swimmer during experiments, we uniformly attached four infrared reflective markers along its body: one at head, two at joints, and one at tail. An OptiTrack motion capture system with six OptiTrack Flex 13 cameras was used to track the three-dimensional positions of the markers at a frame rate of 120 FPS. We recorded trajectories of swimmer movement and calculated displacements and body shape patterns to evaluate locomotor performance.

In each experiment, the swimmer was positioned such that its entire body was fully immersed in the granular medium, with the top edge maintained at a depth of 1 cm beneath the surface. For every gait tested, the swimmer executed seven full cycles. To eliminate the effects of transient dynamics, only the final five cycles were included in the analysis. Each experiment was repeated five times, and we report averaged results with standard deviation across trials. Additionally, we conducted friction characterization tests by dragging the swimmer in a static configuration (i.e., without any active joint motion) through the granular bed. These tests provided baseline measurements of resistive forces and supported interpretation of locomotion performance.

\section{Results}

\subsection{Verification of system dynamics for shape prediction with body compliance}
A critical first step in validating our framework is to confirm that the system dynamics model (Eq.~\ref{eq:dyn}) correctly predicts the actual body shapes that emerge when motor commands are filtered through compliance and environmental interactions. To test this, we prescribed a nominal circular gait in motor space, following the gait equation 
\begin{equation}
\begin{aligned}
\psi_1(t) &= A_\psi\cos(2\pi\omega t),\\
\psi_2(t) &= A_\psi\sin(2\pi\omega t).
\label{eq:serpenoid}
\end{aligned}
\end{equation}
Specifically, we selected $A_\psi = \pi/3, \omega =0.1$ Hz (one gait cycle takes 10 sec). We then examined the resulting trajectories in shape space under different levels of generalized compliance $G$. Representative cases are shown in Fig.~\ref{fig:G}.

With no compliance ($G=0$), the swimmer can accurately execute the prescribed gait commands where its trajectory in the shape space ($\alpha_1$-$\alpha_2$ space) forms a perfect circle. At low compliance ($G=0.25$), the emergent gait closely followed the commanded circle, indicating that the motor commands were faithfully transmitted to the joints with little deformation. At moderate compliance ($G=0.75$), the emergent gait deviated substantially from the nominal circular path, reflecting the growing influence of environmental torques relative to joint stiffness. At high compliance ($G=1.25$), the emergent gait collapsed into a much smaller, distorted loop, demonstrating that environmental forces dominate the effective shape dynamics when the joints are highly compliant.

Across all conditions, the predicted shape trajectories generated by our dynamics model (red) showed excellent agreement with experimentally observed trajectories (gray), capturing both the degree and direction of gait deformation as compliance increased. This agreement demonstrates that the system dynamics model accurately encodes the interaction between motor commands, joint compliance, and environmental resistance. In effect, the model provides a quantitative description of how compliance reshapes commanded motor gaits into realized body motions in shape space.

\begin{figure}[t]
\centering
\includegraphics[scale=.9]{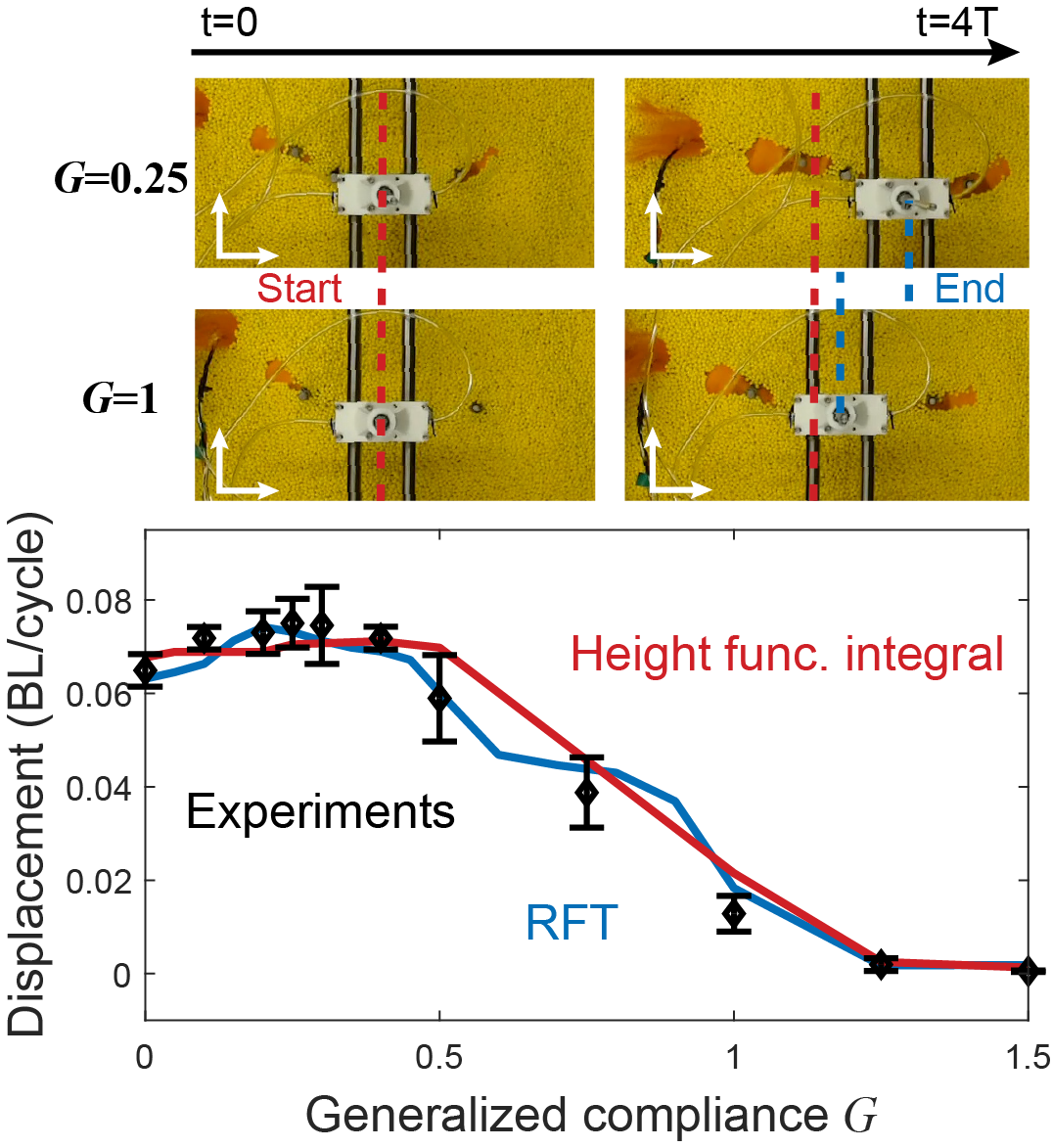}
\caption{\textbf{Verification of locomotor performance predictions under varied body compliance.} (Top) Experimental snapshots of swimmer displacement after four gait cycles for moderate compliance ($G=0.25$) and high compliance ($G=1$). Red and blue dashed lines indicate start and end positions, respectively. At higher compliance, displacement per cycle decreases markedly. (Bottom) Quantitative comparison of displacement per cycle (in body lengths, BL) as a function of generalized compliance $G$. Predictions from resistive force theory (RFT, blue) and geometric mechanics height-function integral (red) closely match experimental measurements (black points with error bars representing standard deviation), capturing the initial plateau at low compliance and sharp performance drop at high compliance.}
\label{fig:Velo}
\end{figure}

\begin{figure}[t]
\centering
\includegraphics[scale=.9]{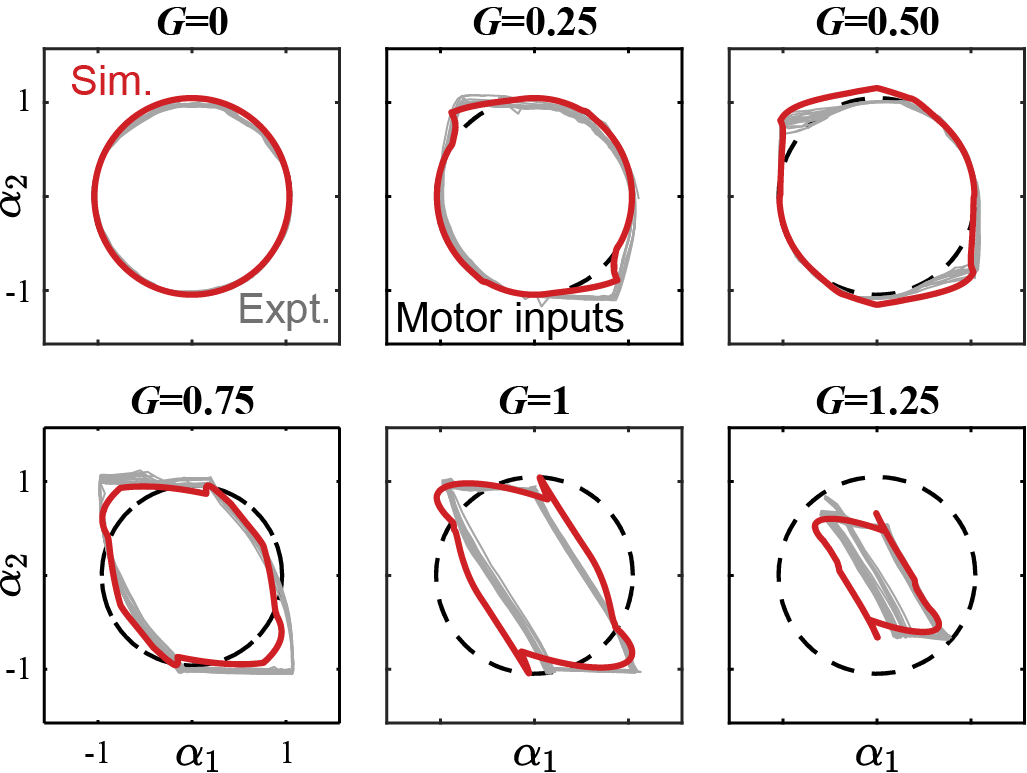}
\caption{\textbf{Validation of system dynamics for predicting emergent shapes under compliance.}
Emergent gait trajectories in shape space ($\alpha_1$-$\alpha_2$) for increasing generalized compliance $G$. Dashed circles indicate prescribed circular motor inputs, gray lines show experimental measurements, and red lines denote simulation predictions. With no compliance ($G=0$), emergent gaits closely follow the commanded circular inputs, while body compliance causes distorted and collapsed trajectories. Simulation results capture the deformation trends observed in experiments across all compliance regimes.}
\label{fig:shape}
\end{figure}

\subsection{Verification of the full geometric mechanics model for locomotor performance prediction}
While the dynamics model captures how commanded gaits are transformed into emergent trajectories, locomotor performance depends on how these trajectories interact with environmental forces to generate net displacement. To assess whether our framework can also predict performance outcomes, we quantified the displacement per cycle as a function of generalized compliance $G$ (Fig.~\ref{fig:Velo}).

Two complementary theoretical tools were employed. First, resistive force theory (RFT, blue) was used to compute the net forces and torques acting on the body as it executed the realized gait, providing a direct estimate of displacement. Second, geometric mechanics was used to derive the height function (red), which encodes the geometric phase associated with a closed trajectory in shape space and predicts the net displacement resulting from each gait cycle.

The relationship between compliance and performance was not strictly monotonic. In the low-compliance regime ($G\approx0.25$), performance was nearly unchanged compared to the non-compliant case and in some conditions even slightly higher. This occurred because environmental perturbations naturally distorted the commanded circular gait in a way that favored forward locomotion; in terms of the height function, the deformed trajectory enclosed more positive area, which increased net displacement. At higher compliance levels, however, performance declined sharply. As gait amplitude collapsed, the enclosed area in the height function shrank, leading to reduced displacement per cycle.

Experimental measurements (black) closely matched both the RFT and height-function predictions, reproducing the initial plateau or slight improvement at low compliance and the drop-off at higher compliance. These results confirm that our framework predicts not only the deformation of gait trajectories but also their performance consequences. They also show that a moderate amount of compliance can sometimes improve locomotion, while excessive compliance ultimately reduces effectiveness in viscous fluid and granular media. In particular, Fig.~\ref{fig:shape} demonstrates that the system dynamics model captures the detailed deformation of gait trajectories across compliance regimes, with predicted emergent shapes (red) aligning closely with experimentally measured ones (gray). This agreement underscores the ability of our framework to predict how motor commands generates emergent body shapes under compliance in granular media.

\subsection{Gait optimization with the proposed framework}
While circular-input gaits provided a convenient baseline to test our model, they did not achieve optimal locomotor performance across different compliance levels. In particular, performance declined sharply as compliance increased, even though the underlying system dynamics were accurately predicted. To address this limitation, we employed our proposed optimization framework, which combines system dynamics, resistive force theory, and geometric mechanics to search for the input (commanding) gaits that maximize displacement given the swimmer's compliance.

\begin{figure}[t]
\centering
\includegraphics[width=0.8\columnwidth]{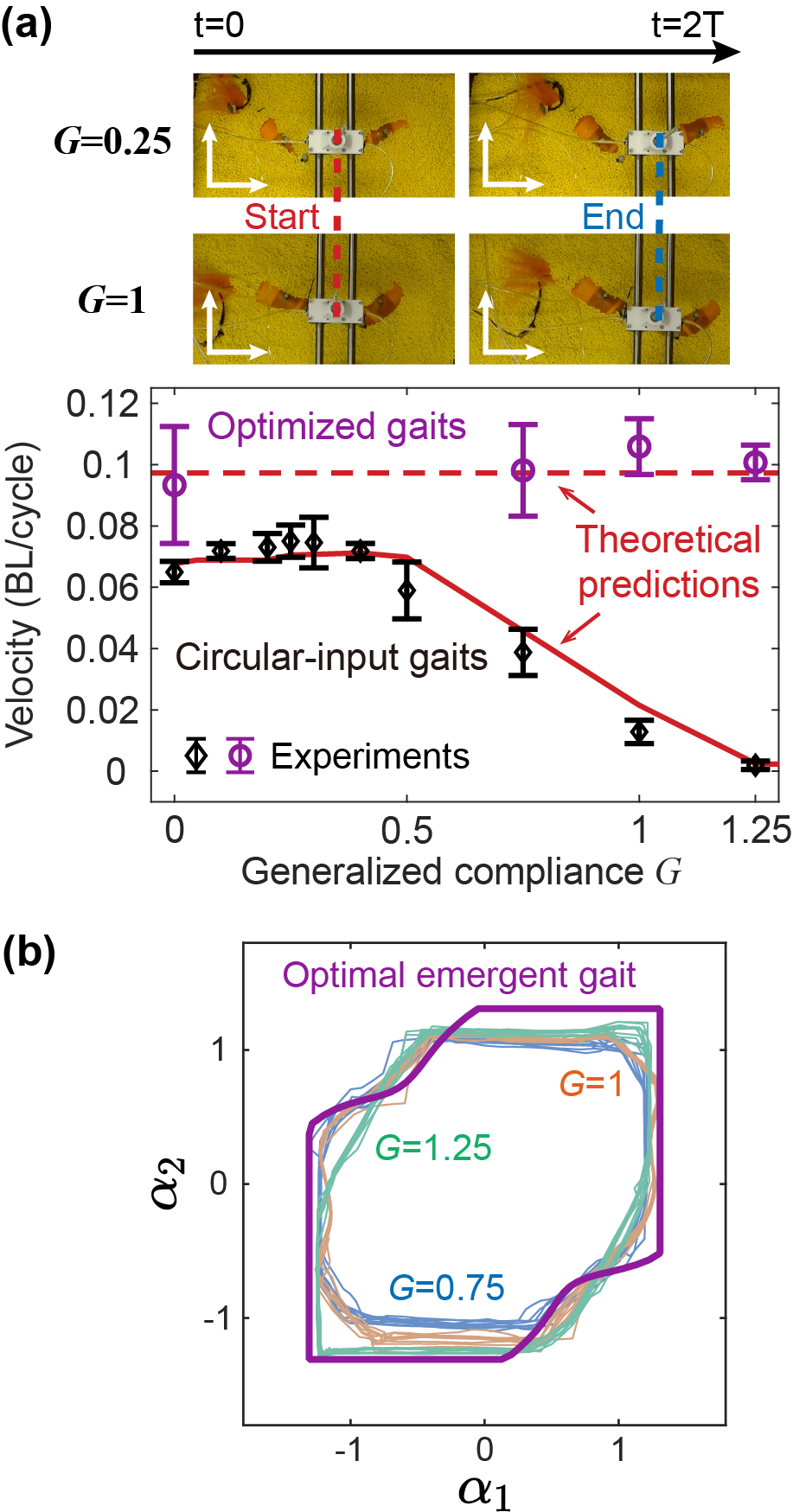}
\caption{\textbf{Gait optimization maximizes performance in compliant swimmers.} (a) Displacement per cycle as a function of generalized compliance $G$. Circular-input gaits (black points) show rapid performance degradation with increasing compliance, consistent with geometric mechanics predictions (red). Optimized gaits (purple points), identified through the proposed optimization framework, maintain consistently high performance across all compliance levels. (b) Emergent gait trajectories in shape space for high compliance levels ($G=0.75,1,1.25$). Despite variations in compliance, the optimized motor inputs yield emergent gaits that closely match the theoretical optimal trajectory (purple), enabling robust high performance.}
\label{fig:optiR_1}
\end{figure}

\begin{figure}[t]
\centering
\includegraphics[width=0.8\columnwidth]{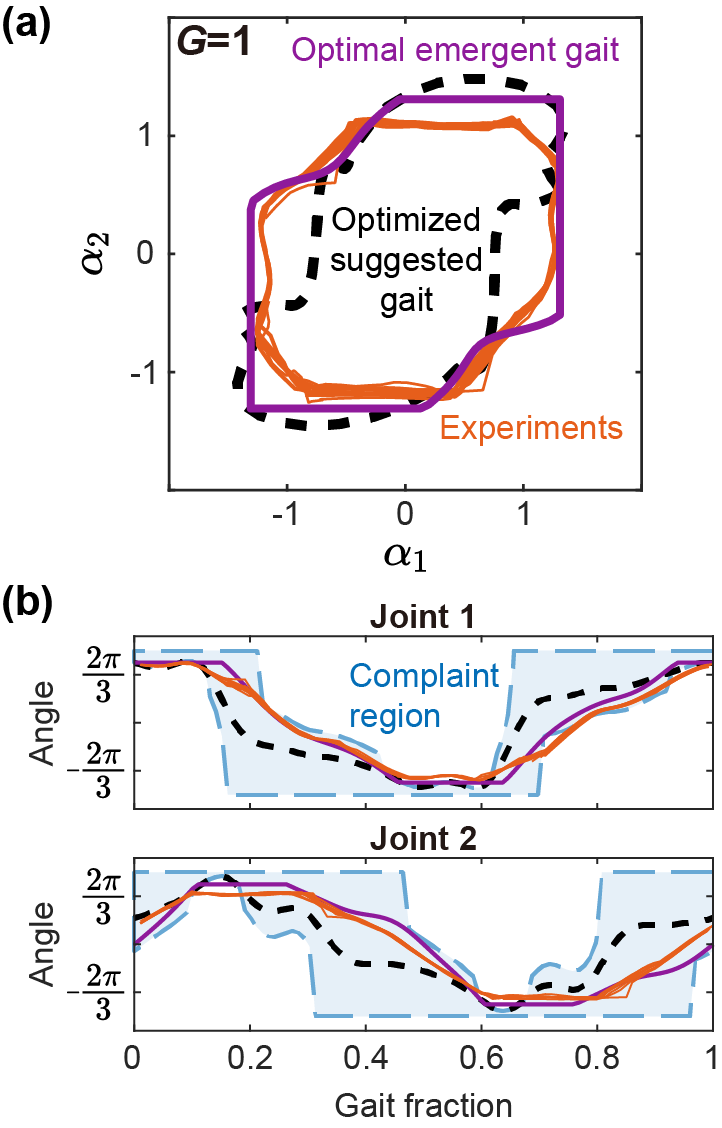}
\caption{\textbf{Execution of optimized gaits in compliant swimmers.} (a) Shape-space trajectories $G=1$. The optimized suggested gait (black dashed) differs substantially from the theoretical optimal emergent gait (purple), and by executing the optimized suggested gait, the swimmer can successfully realize the optimal emergent gait (orange). (b) Joint angle traces over one gait cycle. The emergent joint angles (orange) follow the optimal trajectories (purple), demonstrating that environmental perturbations and compliance are harnessed to reinforce performance.}
\label{fig:optiR_2}
\end{figure}

Overall, Fig.~\ref{fig:optiR_1}(a) compares the displacement of circular suggested gaits (black points, red curve) with that of optimized suggested gaits (purple points). Circular gaits exhibited performance that was high only in the rigid or near-rigid regime and degraded rapidly with increasing compliance. In contrast, optimized suggested gaits maintained consistently high performance, nearly doubling the displacement relative to circular gaits in the compliant regime. Theoretical predictions and experimental measurements were in close agreement, confirming that the optimization framework successfully identifies gaits that recover performance otherwise lost to compliance.

In detail, we first identified the theoretical optimal emergent gait using the RFT and geometric mechanics models, and then applied the inverse system dynamics to determine the corresponding suggested motor commands for each compliance value $G$. Deploying these optimized suggested gaits on the swimmer enabled it to realize the optimal emergent gait and maintain high performance. Fig.~\ref{fig:optiR_1}(b) presents a collection of emergent gait trajectories across different $G$ values. Taking $G=1$ as an example, the optimized suggested gait (black dashed) differs substantially from the circular input, and the swimmer reliably executed the optimal emergent gait trajectory in experiments (Fig.~\ref{fig:optiR_2}(a)). Joint angle traces further confirm that the optimized suggested gait effectively leverages compliant regions of the dynamics (Fig.~\ref{fig:optiR_2}(b)), shaping motor inputs so that environmental perturbations naturally reinforce forward locomotion.

Together, these results demonstrate that optimization does not simply ``fix" the limitations introduced by compliance but instead co-designs gaits with compliance. Without optimization, compliance might appear as a disadvantage, since naive circular gaits collapse and lose performance. With optimization, however, compliance becomes a resource. Unlike rigid robots, which transmit environmental disturbances directly to their actuators and therefore require complex sensing and control to remain stable, compliant robots can passively absorb shocks, adapt their body shape when moving through granular media or clutter, and reduce the burden on active control. In this way, compliance provides a built-in form of mechanical intelligence that improves robustness across environments. Beyond robotics, incorporating compliance also increases the biological relevance of our robophysical models, since animals rely on body elasticity for effective locomotion. Thus, optimization enables compliant robots to achieve high performance while simultaneously establishing them as powerful physical models for studying the role of compliance in biological systems.

\subsection{Optimized compliance for robust obstacle navigation}

\begin{figure*}[t]
\centering
\includegraphics[width=1\textwidth]{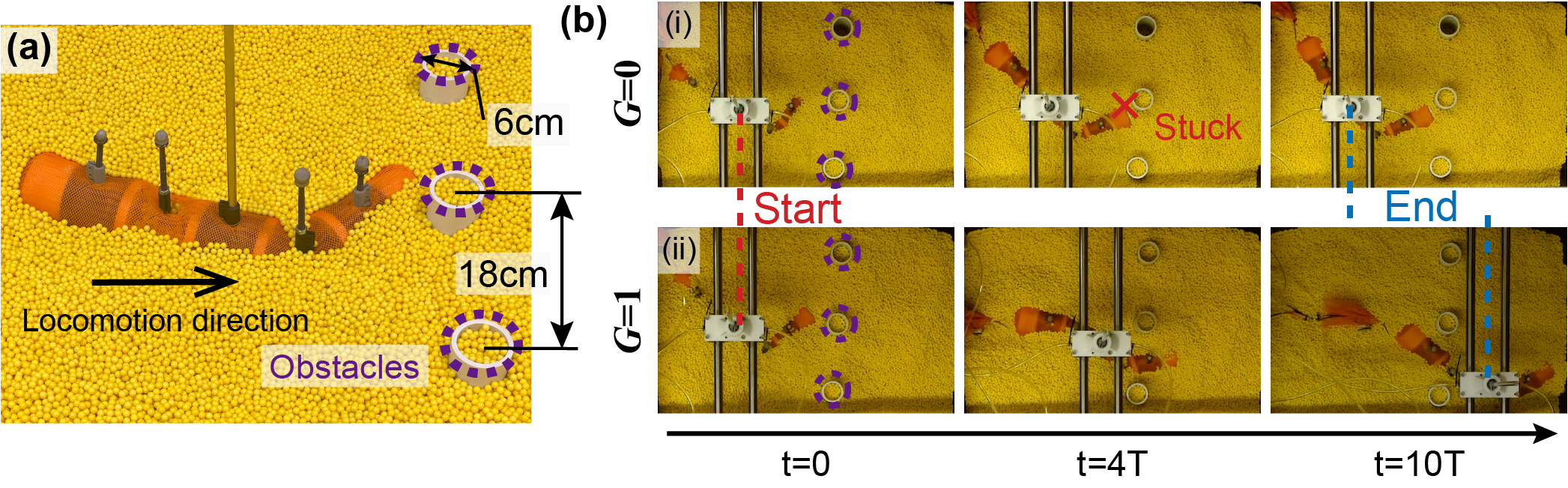}
\caption{\textbf{An optimized gait with body compliance enables the robot to traverse obstacles.} (a) Three cylindrical rigid obstacles immersed in the granular medium. (b.i) When operating the optimized gait without body compliance ($G = 0$), the robot becomes stuck by obstacles in the granular medium. (b.ii) With body compliance ($G = 1$), the robot executing the optimized gait successfully passes through the obstacles while maintaining the speed.}
\label{fig:obstacles}
\end{figure*}

While the previous sections demonstrated how compliance reshapes motor commands and how optimization can recover performance in homogeneous environments, the true advantage of compliance becomes evident in cluttered settings. To test this, we introduced a row of rigid pillars embedded in granular media (Fig.~\ref{fig:obstacles}(a)), where the swimmer was required to pass through the openings between pillars.

\begin{figure}[t]
\centering
\includegraphics[width=1\columnwidth]{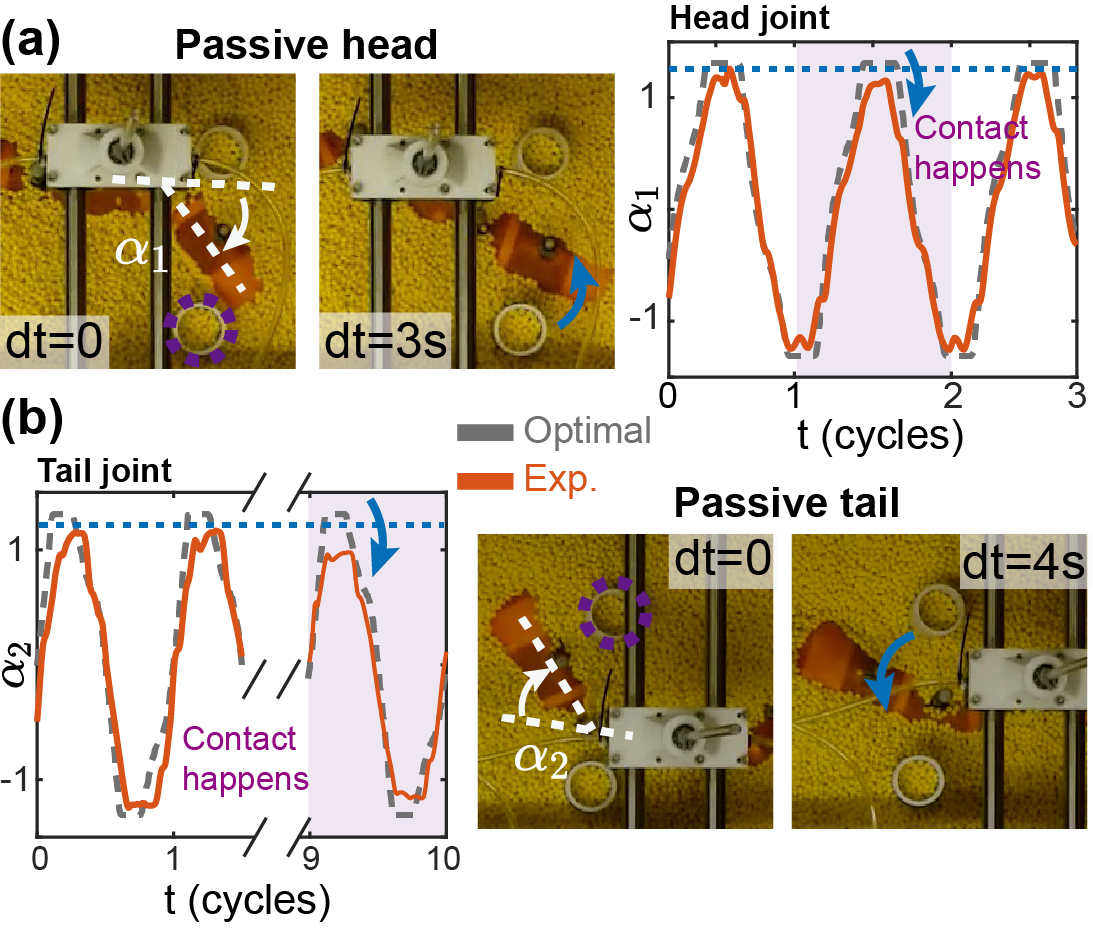}
\caption{\textbf{Passive responses from compliance augment locomotion capabilities.} (a) Compliance in the head joint enables a passive hooking response that aids locomotion upon contact with obstacles. (b) Compliance in the tail joint allows passive deflection, reducing resistance from obstacles.
}
\label{fig:zoomobstacles}
\end{figure}

When operated with no compliance ($G = 0$), the swimmer jammed upon its contacts with the obstacles as shown in Fig.~\ref{fig:obstacles}(b.i). With a rigid body configuration, the robot was unable to reconfigure its shape to exploit the available openings. In some cases, external contact forces were transmitted directly to the actuators, potentially driving them toward torque limits and further preventing forward progression. By contrast, with high compliance ($G = 1$), the swimmer was able to negotiate the obstacles smoothly as illustrated in Fig.~\ref{fig:obstacles}(b.ii). Passive joint flexibility allowed the body to bend around the pillars and ``flow" through the available gaps without requiring explicit sensing or contact modeling. To further explain the role of body compliance in facilitating obstacle traversal, we examined the joint-level responses of the robot executing the optimized gait. Fig.~\ref{fig:zoomobstacles}(a) shows at the head, compliance allows the joint to adapt passively upon contact with an obstacle, producing a hooking response that effectively leverages the obstacle to aid forward motion. This behavior is reflected in the joint-angle trajectory, which deviates from the prescribed optimal gait during contact. At the tail (Fig.~\ref{fig:zoomobstacles}(b)), compliance enables passive deflection as the segment encounters an obstacle, thereby reducing resistive forces that would otherwise impede motion.

It is worth noting that compliance alone is not sufficient for obstacle navigation. Without gait optimization, the compliant swimmer could not achieve effective forward speed in homogeneous media, and thus would also fail to make useful progress in cluttered environments. Our optimization framework resolves this limitation by identifying motor commands that maximize displacement with varied $G$. With optimized gaits, the swimmer maintains high free-space speed, and compliance then naturally takes over the task of negotiating unpredictable obstacle contacts.

This result illustrates a practical control principle: in real-world environments, it is impossible to explicitly model every unknown contact. Instead, robots should be optimized for performance with compliance in homogeneous media, and then rely on compliance to handle unmodeled obstacle interactions. This approach exemplifies the philosophy of mechanical intelligence in robotics, in which passive body mechanics are deliberately leveraged to offload control and enhance robustness. In this way, compliance transforms from a performance liability into a robust locomotor strategy, allowing swimmers to move effectively through clutter where purely rigid robots fail.

\section{Conclusion}

In this work, we developed and experimentally validated a unified framework for modeling, analyzing, and optimizing the locomotion of compliant cable-driven limbless swimmers in highly damped environments. Starting from Purcell’s three-link swimmer, we introduced body compliance by incorporating series elastic elements into the system dynamics, which allowed us to map from suggested joint angles (input body shape sequence) to emergent joint angles (realized body shape sequence). Coupled with resistive force theory and geometric mechanics, this framework enabled us to predict how compliance reshapes commanded gaits and how these emergent trajectories interact with the environment to generate locomotion. Through robophysical experiments with a three-link swimmer in granular media, we verified that the framework accurately predicts both gait deformation and locomotor performance. Finally, we applied the optimization pipeline to identify gaits that maximize displacement under different levels of compliance and showed that swimmers executing these optimized gaits achieve consistently high performance across compliance regimes. Our results extend the gait-design framework of geometric mechanics from systems with prescribed shape control to those in which shape emerges from interactions between the environment and body elasticity. Furthermore, geometric mechanics plays a vital role in simplifying the search for optimal control strategies.

Beyond these, this study provides deeper insight into the role of compliance in limbless locomotion. Our findings reveal that, beyond its well-recognized benefits for interacting with environmental heterogeneity, compliance should not be viewed solely as a constraint or deficiency even in homogeneous environments. While naive circular gaits degrade with increasing compliance, optimized gaits exploit compliance to enhance robustness and adaptability. Unlike rigid robots, which transmit external disturbances directly to actuators and therefore require sophisticated sensing and control to maintain stability, compliant robots can passively absorb shocks, adapt their body shape to cluttered or granular terrain, and reduce control complexity. This passive adaptability represents a form of mechanical intelligence that allows robots to offload part of the locomotor ``control problem" to their body mechanics. This work further contributes to showing how compliance can be implemented while still achieving optimal performance, allowing compliance to become a free bonus rather than a drawback in homogeneous environments. Moreover, embedding compliance into robotic platforms strengthens their connection to biology, since animals universally rely on body elasticity for effective movement.

This work opens up several new directions for advancing the study of compliant undulatory locomotion. First, the framework can be extended to more diverse robot morphologies, such as swimmers with longer bodies or additional joints, and to more complex three-dimensional gaits that better capture the richness of animal locomotion. Second, while this study focused on displacement performance, future efforts should investigate the energetic consequences of compliance, examining whether compliant bodies enable more energy-efficient locomotion by reducing actuation effort or exploiting passive body dynamics. Third, by applying this framework in conjunction with comparative biological studies, compliant robophysical models could serve as powerful experimental platforms for testing biomechanical hypotheses and exploring how animals leverage compliance for robustness and efficiency. Together, these directions highlight the potential of compliance-aware geometric mechanics not only to advance the design of resilient robots but also to deepen our understanding of limbless locomotion in the natural world.

\begin{acknowledgments}
\jlin{The authors thank Ross L. Hatton for helpful discussion.} The authors are grateful for funding from Army Research Office Grants W911NF-11-1-0514 and W911NF-20-1-0129. The authors acknowledge the use of ChatGPT (OpenAI) for language editing.

\end{acknowledgments}

\bibliography{references}

\end{document}